\documentclass{article}

\usepackage{spconf,amsmath,epsfig}

\usepackage{times}
\usepackage{epsfig}
\usepackage{graphicx}
\usepackage{amsmath}
\usepackage{amssymb}

\usepackage{subfigure}
\usepackage[table,xcdraw]{xcolor}
\usepackage{multirow}
\usepackage{diagbox}
\usepackage{booktabs}
\usepackage{arydshln}
\usepackage[normalem]{ulem}
\useunder{\uline}{\ul}{}
\usepackage{hyperref}
\usepackage[misc]{ifsym}

\let\OLDthebibliography\thebibliography
\renewcommand\thebibliography[1]{
  \OLDthebibliography{#1}
  \setlength{\parskip}{0pt}
  \setlength{\itemsep}{0pt plus 0.3ex}
}

\begin{document}\sloppy

\def\x{{\mathbf x}}
\def\L{{\cal L}}

\title{Rank-Based Filter Pruning for Real-Time UAV Tracking}
%

\name{Xucheng Wang$^{\dagger *}$\quad  Dan Zeng$^{\ddagger *}$ \quad  Qijun Zhao$^{\S}$\quad  Shuiwang Li$^{\dagger\textrm\Letter}$\thanks{$^*$ These authors contributed equally. \textrm\Letter Corresponding author.} }
\address{$^\dagger$Guilin University of Technology,  $^\ddagger$Southern University of Science and Technology, $^{\S}$Sichuan University\\
xcwang@glut.edu.cn, zengd@sustech.edu.cn, qjzhao@scu.edu.cn, lishuiwang0721@163.com}

\maketitle

\begin{abstract}

Unmanned aerial vehicle (UAV) tracking has wide potential applications in such as agriculture, navigation, and public security. However, the limitations of computing resources, battery capacity, and maximum load of UAV hinder the deployment of deep learning-based tracking algorithms on UAV. Consequently, discriminative correlation filters (DCF) trackers stand out in the UAV tracking community because of their high efficiency. However, their precision is usually much lower than trackers based on deep learning. Model compression is a promising way to narrow the gap~(i.e., effciency, precision) between DCF- and deep learning- based trackers, which has not caught much attention in UAV tracking. In this paper, 
we propose the P-SiamFC++ tracker, which is the first to use rank-based filter pruning to compress the  SiamFC++ model, achieving a remarkable balance between efficiency and precision. Our method is general and may encourage further studies on UAV tracking with model compression. Extensive experiments on four UAV benchmarks, including UAV123@10fps, DTB70, UAVDT and Vistrone2018, show that P-SiamFC++ tracker significantly outperforms state-of-the-art UAV tracking methods.

\end{abstract}
\section{Introduction}
\label{sec:intro}

Unmanned aerial vehicle (UAV) tracking has emerged as a new task and has attracted increasing interest in visual tracking with the promotion and application of UAVs in various fields recently. It has wide potential applications in such as agriculture, navigation, transportation, public security and disaster response \cite{li2020autotrack,cao2021hift}. However, UAV tracking faces more onerous challenges than in general scenes. On the one hand, motion blur, severe occlusion, extreme viewing angle, and scale changes have greatly challenged the \emph{precision} of the UAV tracking algorithms;
on the other hand, limited computing resources, low power consumption requirements, battery capacity limitations, and UAV's maximum load  pose a big challenge  to its \emph{efficiency} as well \cite{li2021learning}. 

At the current level of technology, efficiency is a fundamental problem of UAV tracking. Therefore, discriminative correlation filters (DCF) trackers are usually preferred instead of trackers based on deep learning\cite{li2020autotrack, huang2019learning}. Although tracking precisions of DCF-based trackers have been greatly improved, they are still not comparable to  most state-of-the-art deep learning-based trackers. Very recently, Cao et al. \cite{cao2021hift} proposed an efficient and effective deep tracker for UAV tracking, which used a lightweight backbone for consideration of efficiency and utilized a hierarchical feature transformer to achieve interactive feature fusion of shallow layers and deep layers for robust representation learning \cite{cao2021hift}. This tracker has achieved a  good balance between efficiency and precision and demonstrated state-of-the-art performance in UAV tracking, which suggests that an effective lightweight deep learning-based tracker may be a good alternative to a DCF-based tracker. We are inspired to exploit model compression to narrow the gap between DCF- and deep learning- based UAV trackers.

Model compression is a technique usually used to deploy state-of-the-art deep networks in low-power and resource-constrained edge devices without compromising much on the model's accuracy \cite{choudhary2020comprehensive}. Popular and widely studied methods for model compression \cite{wang2021emerging} include pruning, quantization, low-rank approximation, knowledge distillation, and etc. In this paper, we exploit the rank-based filter pruning method proposed in \cite{lin2020hrank}, which is efficient and straightforward, to compress the SiamFC++, and thus we name the proposed method P-SiamFC++. The pruning method we used is very straightforward and training efficiently since it eliminates the need of introducing additional constraints and retaining the model. SiamFC++ is based on the efficient tracker SiamFC and shows state-of-the-art performance in both precision and speed by introducing a regression branch and a center-ness branch. To the best of our knowledge, we are the first to effectively use model compression to narrow the gap between DCF- and deep learning- based trackers in UAV tracking. We achieve a remarkable balance between efficiency and precision compared with existing CPU-based and deep learning-based trackers.
We believe our work provides a fresh perspective for solving UAV tracking and will attract increasing attention in the UAV tracking community on model compression. Our contributions can be summarized as follows:

\begin{itemize}
	\item We are the first to introduce model compression to UAV tracking to the best of our knowledge, which narrows the gap between DCF-based and deep learning-based trackers. Surprisingly, the proposed method can improve both efficiency and tracking precision. The significant yet unexpected increase of the compressed model over the original one may encourage more work on this method. 
	\item We propose the P-SiamFC++ tracker to exploit rank-based filter pruning for compressing the SiamFC++ model, achieving a remarkable balance between tracking efficiency and precision. Our method is general, real-time, and provides a fresh perspective to solve UAV tracking. 
	\item We demonstrate the proposed method on four UAV benchmarks. Experimental results show that the proposed P-SiamFC++ tracker achieves state-of-the-art performance.
\end{itemize}

\section{Related Works}
\subsection{Visual Tracking Methods}
Modern trackers can be roughly divided into two classes: DCF-based trackers and deep learning-based ones. DCF-based trackers start with a minimum output sum of squared error (MOSSE) filter. After that, DCF-based trackers have made great progress in many variants \cite{li2021learning}. Since DCF-based trackers usually use handcrafted features and can be calculated in the Fourier domain, they can achieve competitive performance with high efficiency. This is why they stood out in UAV tracking community as efficiency is a fundamental issue. However, they hardly maintain robustness under challenging conditions because of the poor representation ability of handcrafted features. 

Thanks to the great success of deep learning, its application in visual tracking has proven to be very successful in recent years and it has significantly improved tracking precision and robustness. As one of the pioneering works, SiamFC \cite{bertinetto2016fully} considered visual tracking as a general similarity-learning problem and took advantage of the Siamese network to measure the similarity between target and search image. Since then, many deep learning-based trackers based on Siamese architectures have been proposed. 
Recently, deeper architectures have been explored in such as SiamRPN++ \cite{li2019siamrpn++}, SiamBAN \cite{chen2020siamese} to improve tracking precision further, but the efficiency is sacrificed to a large extent. In contrast, SiamFC++ \cite{xu2020siamfc++} is a simple but powerful framework that proposed an effective quality assessment branch in classification to improve precision. Unfortunately, despite its excellent GPU speed for UAV tracking, it cannot reach real-time speed on CPU (i.e., with a speed of $<$ 30 FPS). In this paper, we aim to increase the efficiency of SiamFC++ for real-time UAV tracking \cite{xu2020siamfc++} using model compression techniques.

\subsection{Model Compression by Pruning}

Pruning is a common technique to compress a neural network. In general, the pruning structure can be divided into two types: weight pruning and filter pruning. The former involves removing individual weights or neurons, which, however, is difficult to achieve acceleration on general-purpose hardware \cite{blalock2020state}. While the latter involves removing the entire channels or filters, it is much easier to achieve considerable acceleration \cite{lin2020hrank}. Pruning ratios indicate how many weights to remove, and there are usually two ways to determine the ratio or ratios. The first is to pre-define a global ratio or multiple layer-wise ratios.
The second is to adjust the pruning ratio indirectly, for example, with regularization-based pruning methods. However, it demands much engineering tuning to achieve a specific ratio. Pruning criterion determines which weights to prune. For filter pruning, Frobenius norm or sparsity of the filter response, and the scaling factor of the Batch Normalization layer are frequently used criteria \cite{wang2021emerging}. Pruning schedule specifies how the sparsity of the network goes from zero to a target number, which has two typical choices \cite{wang2021emerging}: (1) in a single step (one-shot), then finetune, (2) progressively, pruning and training are interleaved. Although the progressive manner may be better than the one-shot way as there is more time for training, the latter is more efficient in training and can free one from designing complex training strategies.

Overall, pruning remains an open problem so far. Recently, an effective and efficient filter pruning method was proposed in \cite{lin2020hrank}.
It uses the rank of the feature map in each layer as the pruning criterion and is scheduled in an one-shot way without introducing additional constraints or retraining, which greatly simplifies the complexity of pruning. In this work, we utilize this approach to achieve a model compression target for UAV tracking, which has not been well explored so far.

\section{P-SIAMFC++: Pruned Siamese Tracker for UAV Tracking}

\begin{figure*}[h]
	\centering
	\includegraphics[width=1\textwidth,height=0.21\textwidth]{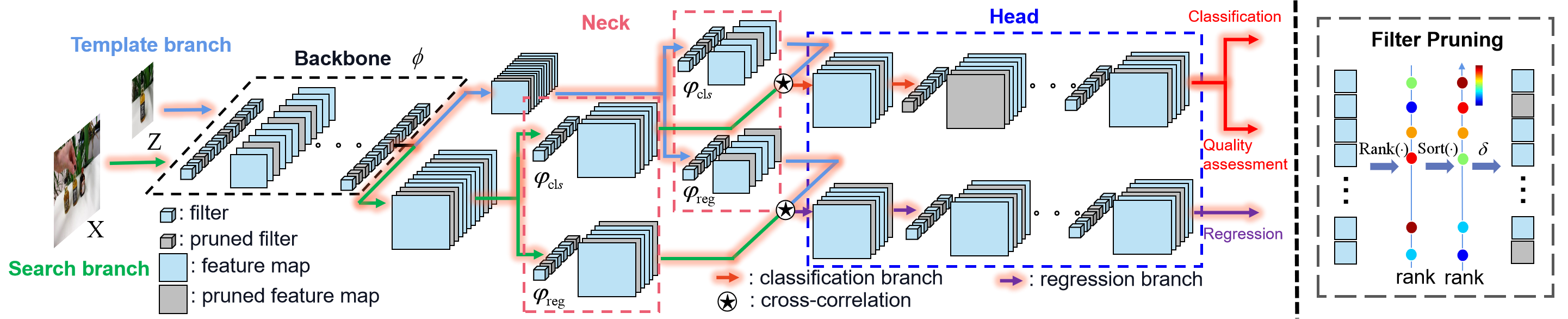}
	\caption{ An illustration of the proposed P-SiamFC++ method. The network structure is inherited from that of SiamFC++. The
difference lies in the pruned filters and feature maps determined by filter pruning (on the right). For simplicity, the subsequent
architectures connected to the head are omitted here since no pruning is involved.
} \label{P-saimfc++_overview}
\end{figure*}

\subsection{P-SiamFC++ Overview}
As illustrated in Fig. \ref{P-saimfc++_overview}, our P-SiamFC++ consists of a template branch, a search branch and three parts: backbone, neck and head. The two branches share the same backbone (we are focusing on Alexnet ) for feature extraction, which is denoted by the mapping $\phi(\cdot)$. The template branch generates features using the tracking target patch Z (as input). The search branch generates features using the search region X. The features of the two branches are coupled with cross-correlation before being used for subsequent classification and regression tasks. The coupled features are defined as follows:

\begin{equation}
\small
    f_l(Z,X) = \psi_l(\phi(Z)) \star \psi_l(\phi(X)), l\in \{cls, reg\},
\end{equation}
where $\star$ denotes the cross-correlation operation, and $\psi_l(\cdot), l\in \{cls, reg\}$ denotes the task-specific layer (`cls' and `reg' are short for classification and regression, respectively). Note that the output of $\psi_{cls}$ and $\psi_{reg}$ are of the same size, and a center-ness branch is in parallel with the classification branch for assessing classification qualities and finally reweighting the classification scores.
The overall loss for training P-SiamFC++ is as follow \cite{xu2020siamfc++}:
\begin{equation}\label{EQ_RRCF}
	\small
	\begin{split}
		L (\{p_z\},q_z,\{t_z\})= \frac{1}{N_{pos}}\sum_{z}( L_{cls}(p_z, p_z^*) + \\\lambda_1 I_{\{p_z^*>0\}}L_{qual}(q_z,q_z^*)+ \lambda_2 I_{\{p_z^*>0\}}L_{reg}(t_z,t_z^*))
	\end{split}
\end{equation}
where $z$ represents a coordinate on a feature map, $p_z$is a prediction while $p_z^*$ is the corresponding target label, $I_{\{\cdot\}}$ is the indicator function, $L_{cls}$, $L_{qual}$ and $L_{reg}$ denote the focal loss, the binary cross entropy loss and the IoU loss for classification, quality assessment and regression, respectively. Refer to \cite{xu2020siamfc++} for detail. $\lambda_1$ and $\lambda_2$ are weight terms to balance the losses, $N_{pos}=\sum_{z}I_{\{p_z^*>0\}}$. Note that if $z$ is considered as a positive sample $p_z^*$ is assigned 1, otherwise 0 if considered as a negative sample. The pipeline of our P-SiamFC++ is basically the same as that of SiamFC++. The difference lies in the pruned filters determined by filter pruning, which will be explained in detail in the following subsection.

\subsection{Rank-based Filter Pruning Criterion}
Denote the $i$-th ($i\in [1,K]$) convolutional layer $C^i$ of the SiamFC++ by a set of 3-D filters $W_{C^i}=\{w_1^i,w_2^i,...,w_m^i\}\in \mathbb{R}^{n_i\times n_{i-1}\times k_i \times k_i}$, where $n_i$ denotes the number of filters in $C^i$, $k_i$ is the kernel size, and the $j$-th filter is $w_j^i\in \mathbb{R}^{n_{i-1}\times k_i \times k_i}$. The output feature maps of the filters are denoted by $O_{C^i}=\{o_1^i,o_2^i,...,o_m^i\}\in \mathbb{R}^{n_i\times g\times h_i \times w_i}$, where $o_j^i\in \mathbb{R}^{g\times h_i \times w_i}$ is generated by $w_j^i$, $g$ is the number of input images, $h_i$ and $w_i$ denote the height and width of the feature maps, respectively. The rank-based pruning criterion in \cite{lin2020hrank} is formulated as the following optimization problem:
\begin{equation}\label{Hrank_obj}
\small
\begin{split}
     \underset{\delta_{i,j}}{min}\sum_{i=1}^{K}\sum_{j=1}^{n_i}\delta_{i,j}\mathbb{E}_{I\sim P(I)}[Rank(o_j^i(I))], s.t \sum_{j=1}^{n_i}\delta_{i,j}=n_{p}^i,
\end{split}
\end{equation}
where $I$ denotes an input image which follows the $P(I)$ distribution, $n_p^i$ represents the number of filters to be pruned in $C^i$. $\delta_{i,j}\in \{0,1\}$ indicates whether the filter $w_j^i$ is pruned, $\delta_{i,j}=1$ if it is, otherwise $\delta_{i,j}=0$. $Rank(\cdot)$ computes the rank of a feature map , which is a measure of information richness.The expectation of ranks generated by a single filter is empirically proved to be robust to the input images\cite{lin2020hrank}, on the ground of which Eq. (\ref{Hrank_obj}) is approximated by 
\begin{equation}\label{Hrank_obj_simplified}
\small
\begin{split}
     \underset{\delta_{i,j}}{min}\sum_{i=1}^{K}\sum_{j=1}^{n_i}\delta_{i,j}\sum_{t=1}^{g}Rank(o_j^i(I_t)),\quad s.t \sum_{j=1}^{n_i}\delta_{i,j}=n_{p}^i,
\end{split}
\end{equation}
where $t$ indexes the input images. Eq. (\ref{Hrank_obj_simplified}) be minimized by pruning $n_p^i$ filters with the least average ranks of feature maps.

\subsection{Rank-based Filter Pruning Schedule}
The pipeline of rank-based pruning is as follows: First, calculate the average rank of the feature map of any filter in each layer to obtain the rank set $\{R^i\}_{i=1}^K=\{\{r_1^i,r_2^i,...,r_{n_i}^i\}\}_{i=1}^K$. Second, each $R^i$ is sorted in decreasing order, ending up with $\bar{R}^i=\{r_{s_1^i}^i,r_{s_2^i}^i,...,r_{{s_{n_i}^i}}^i\}$, where ${s_j^i}$ is the index of th $j$-th top value in $R^i$. Third, we empirically determine the number of pruned filters of each layer $n_{p}^{i}$ in order to prune the SiamFC++ model, and then conduct filter pruning to obtain P-SiamFC++. After pruning, $R^i$ turns to $\hat{R}^i=\{r_{s_1^i}^i,r_{s_2^i}^i,...,r_{{s_{\hat{n}_i}^i}}^i\}$ in which $\hat{n}_i=n_i-n_p^i$. Finally, the filters retained are initialized with the original weights in the trained SiamFC++ model, and then the compressed model P-SiamFC++ is finetuned. 

\section{Experiments}
\begin{figure*}[t]
	\centering
	\subfigure{
		\begin{minipage}[t]{0.24\textwidth}
			\includegraphics[width=1\textwidth,height=0.6\textwidth]{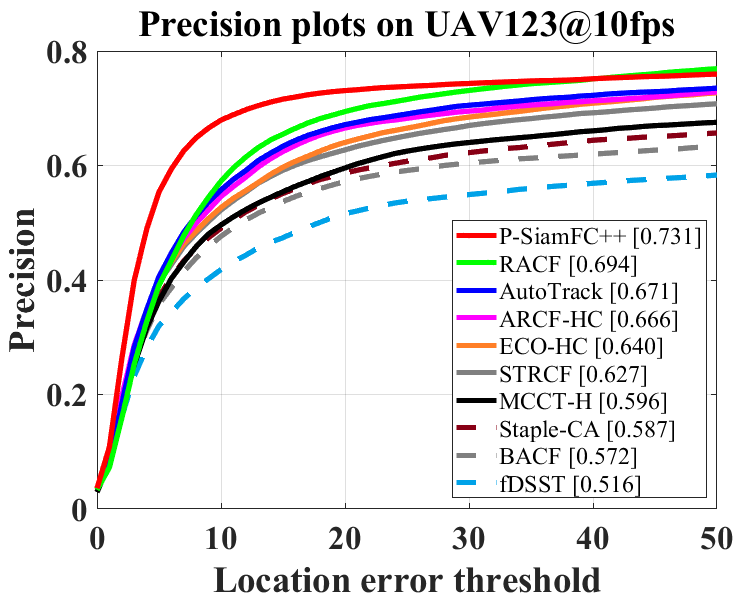}\hspace{0in}
			\includegraphics[width=1\textwidth,height=0.6\textwidth]{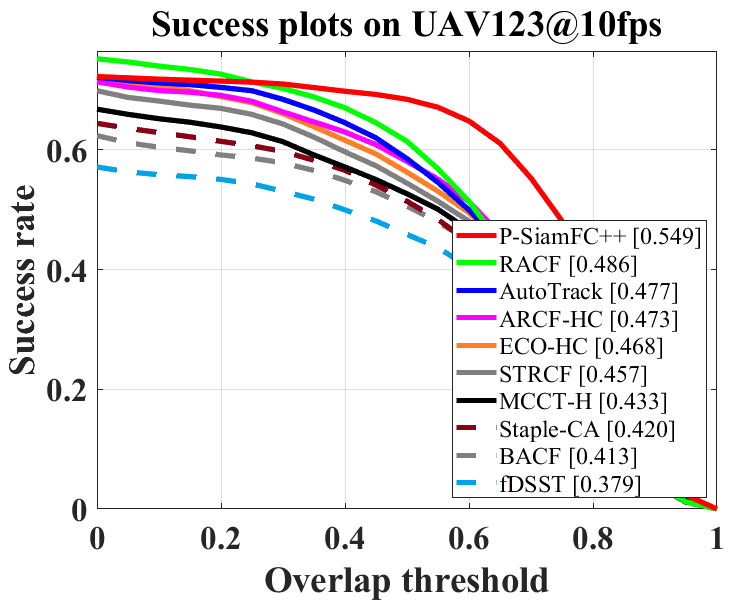}
	\end{minipage}}
	\subfigure{
		\begin{minipage}[t]{0.24\textwidth}
			\includegraphics[width=1\textwidth,height=0.6\textwidth]{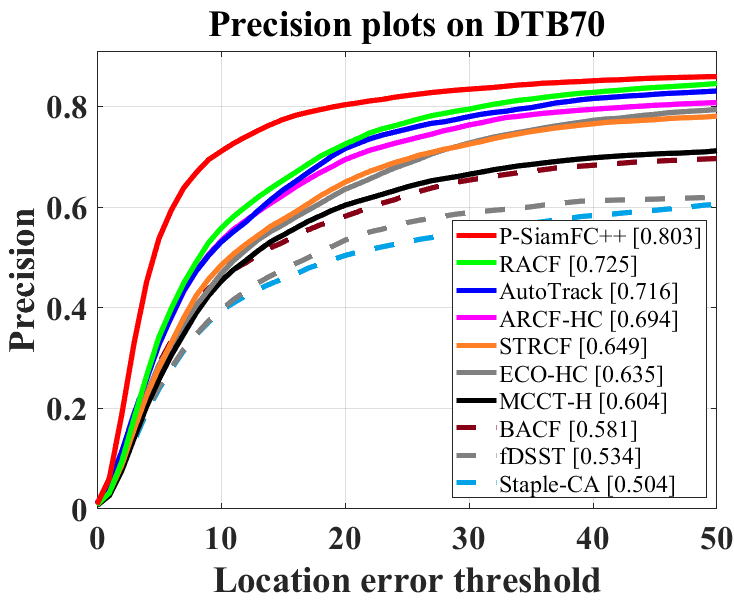}\hspace{0in}
			\includegraphics[width=1\textwidth,height=0.6\textwidth]{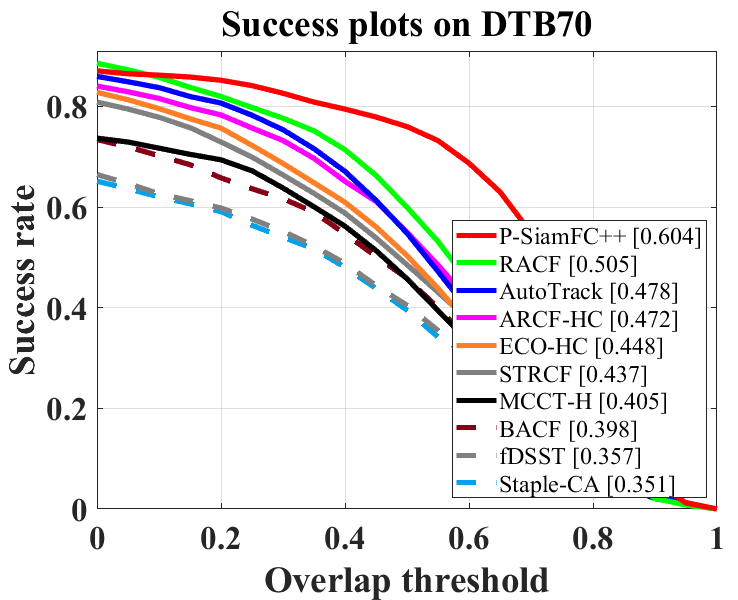}
	\end{minipage}}
	\subfigure{
		\begin{minipage}[t]{0.24\textwidth}
			\includegraphics[width=1\textwidth,height=0.6\textwidth]{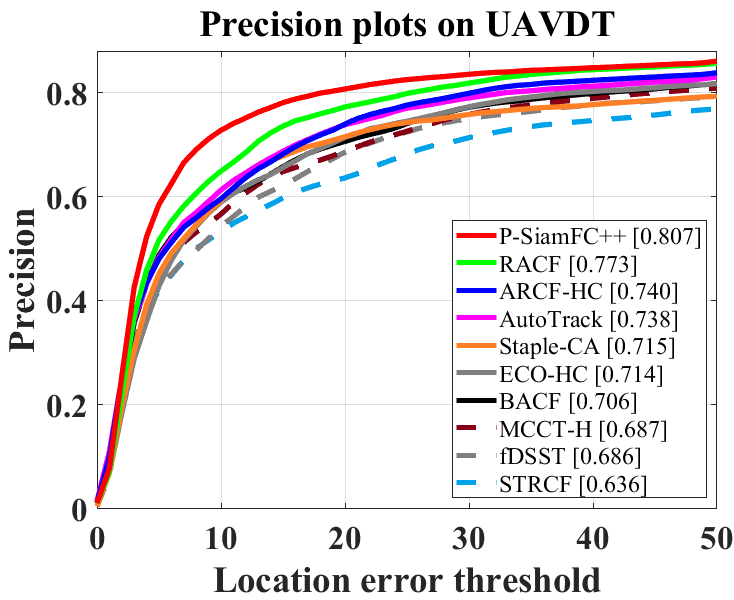}\hspace{0in}
			\includegraphics[width=1\textwidth,height=0.6\textwidth]{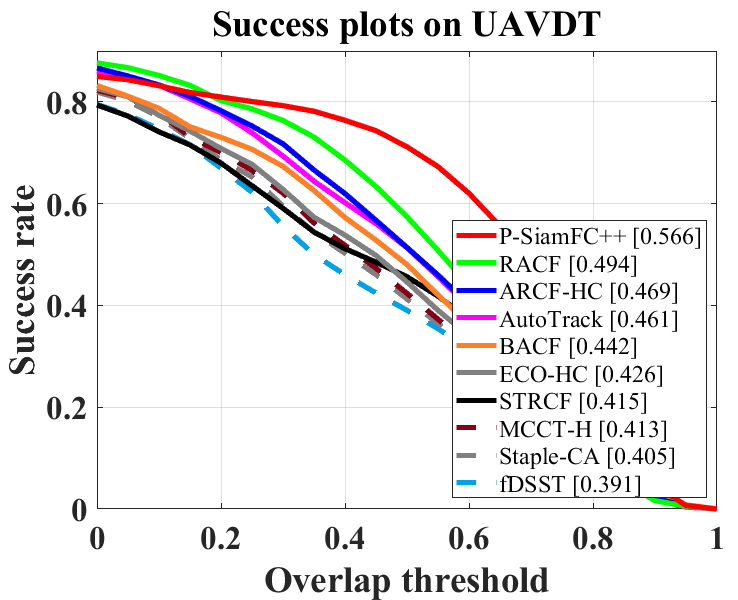}
	\end{minipage}}
	\subfigure{
		\begin{minipage}[t]{0.24\textwidth}
			\includegraphics[width=1\textwidth,height=0.6\textwidth]{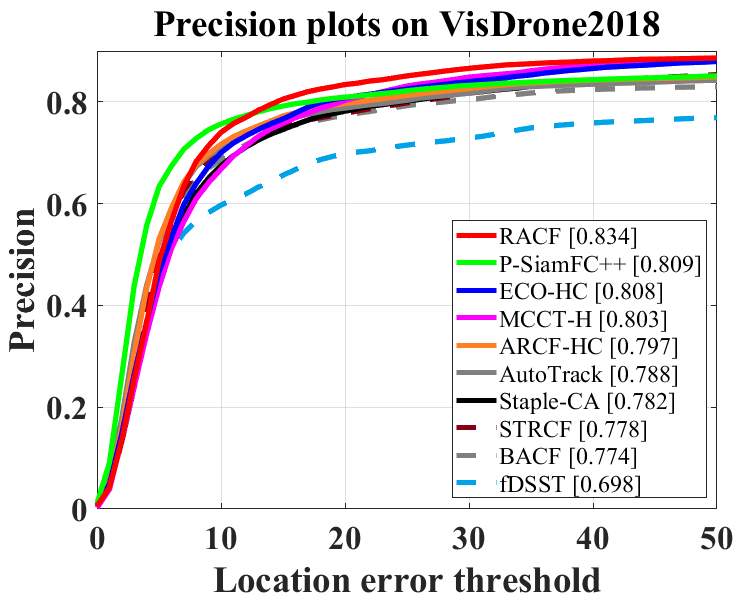}\hspace{0in}
			\includegraphics[width=1\textwidth,height=0.6\textwidth]{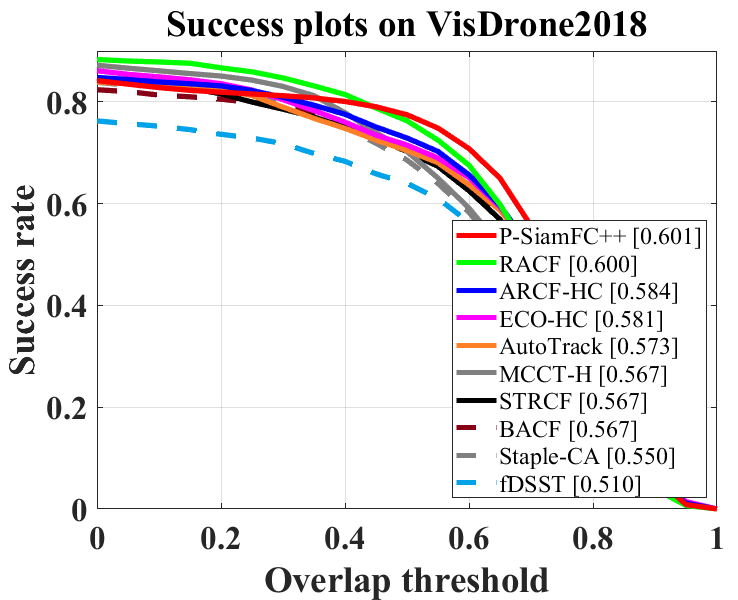}
	\end{minipage}}
	\caption{Overall performance of hand-crafted based trackers on, from left to right, UAV123@10fps, DTB70, UAVDT and VisDrone2018. Precision and success rate for one-pass evaluation (OPE) are used for evaluation. The precision at 20 pixels and area under curve (AUC) are used for ranking and marked in the precision plots and success plots respectively.}
	\label{fig_overall_p_s_plots}
\end{figure*}

\begin{table*}[]
	\footnotesize
	\centering
	\caption{Average precision and speed (FPS) comparision between P-SiamFC++ and hand-crafted based trackers on UAV123@10fps, DTB70, UAVDT and VisDrone2018.  All the reported FPSs are evaluated on a single CPU. {\color[HTML]{FE0000}Red}, {\color[HTML]{3531FF}blue} and {\color[HTML]{009901}green} respectively indicate the first, second and third place.}
	\label{tab:precision_FPS}
	\resizebox{7.0in}{0.285in}{
		\begin{tabular}{@{}cccccccccccccc    ccc@{}}
			\toprule
				& KCF\cite{2015High}& fDSST\cite{danelljan2016adaptive}& Staple-CA\cite{mueller2017context}	& BACF\cite{2017Learning}& ECO-HC\cite{danelljan2017eco}& MCCT-H\cite{wang2018multi}	& STRCF\cite{li2018learning} & ARCF-HC\cite{huang2019learning}  & AutoTrack\cite{li2020autotrack} 
		   &RACF\cite{li2021learning} & \textbf{P-SiamFC++}                                   \\ \hline\hline
			\textbf{Precision} 
			& 53.3& 60.4& 64.2 & 65.3  & 68.8 & 66.8 	 & 67.1   & 71.9 	&\color[HTML]{009901}\textbf{72.3}
				 &  {\color[HTML]{3531FF} \textbf{75.7}}	& {\color[HTML]{FE0000} \textbf{78.8}} 
			\\
			\textbf{FPS (CPU)} & {\color[HTML]{FE0000} \textbf{622.5}}&  {\color[HTML]{3531FF} \textbf{193.4}}& 64.3& 54.2  & {\color[HTML]{009901} \textbf{84.5}}& 63.4 & 28.4 	 & 34.2     & 58.7& 35.7& 46.1        \\ \hline
		\end{tabular}
	}
\end{table*}

\begin{table*}[]
	\caption{Precision and speed (FPS) comparison between P-SiamFC++ and
		deep-based trackers on UAVDT \cite{du2018the}. All the reported FPSs are evaluated on a single GPU. {\color[HTML]{FE0000}Red}, {\color[HTML]{3531FF}blue} and {\color[HTML]{009901}green} indicate the first, second and third place. }
	\label{tab:precision_FPS_deep}
	\resizebox{7.0in}{0.285in}{
		\begin{tabular}{ccccccccccccccc}
			\toprule
			& \begin{tabular}[c]{@{}c@{}}SiamR-CNN\cite{voigtlaender2020siam} \end{tabular} & \begin{tabular}[c]{@{}c@{}}D3S\cite{lukezic2020d3s}\end{tabular} & \begin{tabular}[c]{@{}c@{}}PrDimp18\cite{danelljan2020probabilistic}\end{tabular} & \begin{tabular}[c]{@{}c@{}}KYS\cite{bhat2020know}\end{tabular} & \begin{tabular}[c]{@{}c@{}}SiamGAT\cite{guo2021graph}\end{tabular} & \begin{tabular}[c]{@{}c@{}}LightTrank\cite{yan2021lighttrack}\end{tabular} & \begin{tabular}[c]{@{}c@{}}TransT\cite{chen2021transformer}\end{tabular} & \begin{tabular}[c]{@{}c@{}}HiFT\cite{cao2021hift}\end{tabular} & \begin{tabular}[c]{@{}c@{}}SOAT\cite{zhou2021saliency}\end{tabular} & \begin{tabular}[c]{@{}c@{}}AutoMatch\cite{zhang2021learn}\end{tabular} & \begin{tabular}[c]{@{}c@{}}\textbf{P-SiamFC++} (Ours)\end{tabular} \\ \hline\hline
			\textbf{Precision}  & 66.5                                                        & 72.2                                                  & 73.2                                                       & 79.8                                                  & 76.4                                                      & 80.4                                                         & {\color[HTML]{FE0000} \textbf{82.6}}                     & 65.2                                                   & {\color[HTML]{3531FF} \textbf{82.1}}                   & {\color[HTML]{3531FF} \textbf{82.1}}                        & {\color[HTML]{009901} \textbf{80.7}}                      \\
			\textbf{FPS (GPU)}     & 7.2                                                         & 44.2                                                  & 48.1                                                       & 30.0                                                  & 74.2                                                      & {\color[HTML]{009901} \textbf{84.1}}                         & 41.8                                                     & {\color[HTML]{3531FF} \textbf{134.1}}                  & 29.2                                                   & 50.0                                                        & {\color[HTML]{FE0000} \textbf{258.8}}                       \\ \hline
		\end{tabular}
	}
\end{table*}
\subsection{Experiment settings}
We mainly conduct our experiments on four challenging UAV benchmarks, i.e., UAV123@10fps \cite{2016A}, DTB70 \cite{li2017visual}, UAVDT \cite{du2018the} and Vistrone2018 \cite{wen2018visdrone}. UAV123@10fps is constructed by sampling the UAV123 benchmark from original 30FPS to 10FPS, and is used to study the impact of camera capture speed on tracking performance. DTB70 consists of 70 UAV sequences, which primarily addresses the problem of severe UAV motion, but also includes various cluttered scenes and objects with different sizes. UAVDT is mainly used for vehicle tracking with various weather conditions, flying altitudes and camera views. 
Vistrone2018 is from a single object
tracking challenge held in conjunction with the European conference on computer vision (ECCV2018), which focuses on evaluating tracking algorithms on drones.

All evaluation experiments are conducted on a PC equipped with i9-10850K processor (3.6GHz), 16GB RAM and an NVIDIA TitanX GPU. For the backbone (AlexNet), the pruning ratios of the five convolution layers are (0.792, 0.875, 0.878, 0.870, 1.0). For the head, the pruning ratios of three convolution layers used for classification and regression are (0.898, 0.539, 0.875) and (0.887, 0.566, 0.875), respectively. The neck is not pruned. Other parameters for training and inference follow SiamFC++ \cite{xu2020siamfc++}. Code is available on: \url{https://github.com/P-SiamFCplusplus2021/P-SiamFCplusplus2021}
	
\begin{figure}[h]
	\centering
	\includegraphics[width=0.475\textwidth,height=0.31\textwidth]{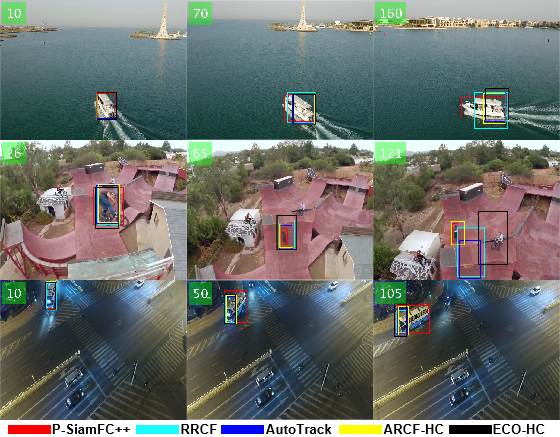}
	\caption{Qualitative evaluation on 3 sequences from, respectively, UAV123@10fps, DTB70 and UAVDT (i.e. boat3, BMX3 and S0304). The results of different methods have been shown with different colors.} 
	\label{fig:visual_examples}
\end{figure}

\subsection{Comparison with CPU-based trackers}
 
Ten state-of-the-art trackers based on hand-crafted features for comparison are: KCF \cite{2015High}, fDSST \cite{danelljan2016adaptive}, Staple-CA \cite{mueller2017context}, BACF \cite{2017Learning}, ECO-HC \cite{danelljan2017eco}, MCCT-H \cite{wang2018multi}, STRCF \cite{li2018learning}, ARCF-HC \cite{huang2019learning}, AutoTrack  \cite{li2020autotrack}, RACF \cite{li2021learning}.

\begin{table*}[t]
	\centering
	\caption{Comparison of model size (parameters), precision and tracking speed between proposed P-SiamFC++ and the baseline method SiamFC++ on four UAV benchmarks. PRC is short for precision. Note that only the precision on CPU is shown here since the difference of precision on CPU and GPU is very small.}
	\label{tab:compareWbaseline}
	\resizebox{6.2in}{0.41in}{
		\begin{tabular}{@{}ccccccccccccccccc@{}}
			\toprule
			\multirow{3}{*}{Methods} & \multirow{3}{*}{Parameters} & \multicolumn{3}{c}{UAV123@10fps}                      & \multicolumn{3}{c}{DTB70}                             & \multicolumn{3}{c}{UAVDT}                             & \multicolumn{3}{c}{VisDrone2018}                      & \multicolumn{3}{c}{Avg.}                              \\ \cmidrule(l){3-5} \cmidrule(l){6-8} \cmidrule(l){9-11} \cmidrule(l){12-14} \cmidrule(l){15-17}
			&                         & \multirow{2}{*}{PRC} & \multicolumn{2}{c}{FPS}        & \multirow{2}{*}{PRC} & \multicolumn{2}{c}{FPS}        & \multirow{2}{*}{PRC} & \multicolumn{2}{c}{FPS}        & \multirow{2}{*}{PRC} & \multicolumn{2}{c}{FPS}        & \multirow{2}{*}{PRC} & \multicolumn{2}{c}{FPS}        \\ \cmidrule(l){4-5} \cmidrule(l){7-8} \cmidrule(l){10-11} \cmidrule(l){13-14} \cmidrule(l){16-17}
			&                         &                      & CPU           & GPU            &                      & CPU           & GPU            &                      & CPU           & GPU            &                      & CPU           & GPU            &                      & CPU           & GPU            \\ \hline \hline
			SiamFC++                 & 9.66M                      & 72.8                 & 29.0          & 212.1          & \textbf{80.5}        & 29.8          & 219.3          & 76.2                 & 30.0          & 238.9          & 72.5                 & 28.9          & 205.8          & 75.5                 & 29.4          & 219.1          \\
			P-SiamFC++ (Ours)              & 7.49M                      & \textbf{73.1}        & \textbf{45.1} & \textbf{236.4} & 80.3                 & \textbf{45.6} & \textbf{238.2} & \textbf{80.7}        & \textbf{48.8} & \textbf{258.8} & \textbf{80.9}        & \textbf{45.0} & \textbf{230.5} & \textbf{78.8}        & \textbf{46.1} & \textbf{241.0} \\ \bottomrule
		\end{tabular}
	}
\end{table*}
\indent\textbf{Quantitative evaluation:} The overall performance of P-SiamFC++ with the competing trackers on the four benchmarks is shown in Fig. \ref{fig_overall_p_s_plots}. It can be seen that P-SiamFC++ outperforms all other trackers on all four benchmarks.
Specifically, on UAV123@10fps, DTB70 and UAVDT, P-SiamFC++ significantly outperforms the second tracker RACF in terms of precision and AUC, with gains of (3.7\%, 6.3\%), (7.8\%, 9.9\%) and (3.4\%, 7.2\%), respectively. On VisDrone2018, P-SiamFC++ is inferior to the first tracker RACF in precision, the gap is 2.5\%. However, we achieve the first place in terms of AUC. Note that the parameters of RACF is dataset specific. In terms of speed, we use the average FPS over the aforementioned four benchmarks on CPU as the metric of tracking. Table \ref{tab:precision_FPS} illustrates average precision and FPS produced by different trackers. As can be seen, P-SiamFC++ outperforms all the competing trackers in precision, and is also the best real-time tracker (speed of $>$30FPS) on CPU. Specifically, P-SiamFC++ achieves 78.8\% in precision at a speed of 46.6 FPS.

\indent\textbf{Qualitative evaluation:}
We show some qualitative tracking results of our method and four top CPU-based trackers in Fig. \ref{fig:visual_examples}. It can be seen that the four CPU-based trackers fail to maintain robustness in challenging examples where objects are experiencing large deformation (i.e., BMX3) or pose change (i.e., boat3 and S0304), but our P-SiamFC++ performs much better and is visually more satisfying by virtue of the deep representation learning. This suggests that developing more efficient deep trakcers for UAV tracking may be more effective in improving tracking precision.

\subsection{Comparison with deep learning-based trackers}
The proposed P-SiamFC++ is also compared with ten state-of-the-art
deep trackers on the UAVDT dataset, including PrDiMP18 \cite{danelljan2020probabilistic}, SiamR-CNN \cite{voigtlaender2020siam}, D3S \cite{lukezic2020d3s}, KYS \cite{bhat2020know}, SiamGAT \cite{guo2021graph}, LightTrack \cite{yan2021lighttrack}, TransT \cite{chen2021transformer}, HiFT \cite{cao2021hift}, SOAT \cite{zhou2021saliency}, AutoMatch \cite{zhang2021learn}. 

The FPSs and the precisions on UAVDT are shown in Table \ref{tab:precision_FPS_deep}. As can be seen, although P-SiamFC++ is inferior in precision to TransT, SOAT and AutoMatch, the gaps are less than 2\% and the GPU speed of P-SiamFC++ is 6 times greater than the first tracker TransT. This justifies that our P-SiamFC++ has achieved a good balance between precision and efficiency (i.e., speed).

\subsection{Ablation study}
\indent\textbf{Effect of pruning:} 
To see how the model size, precision and tracking speed will change when the rank-based filter pruning is applied to the baseline method SiamFC++, we compare the proposed P-SiamFC++ with the baseline SiamFC++ on all the four UAV benchmarks. Their comparison in terms of model size, precisions and speeds are shown in Table \ref{tab:compareWbaseline}. As can be seen, the model size of P-SiamFC++ is reduced to 77.6\% ($\approx$7.49/9.66) of the original. Both the CPU and GPU speed are improved. Since the parallel computing units on our GPU far exceeds the size of both models, the GPU speed has increased by only 11\% on average. But the average CPU speed is raised from 29.4 FPS to 46.1 FPS, an increase of 56.8\%. Remarkably, although P-SiamFC++ is slightly inferior to the baseline SiamFC++ on DTB70, the precision improvement on UAVDT and VisDrone2018 is very significant, specifically, with gains of 4.5\% and 8.4\%, respectively. These results justify that the proposed method is effective in improving both efficiency and precision.
\begin{table}[h]
\centering
\caption{Illustration of how the precision of P-SiamFC++ on DTB70 varies with the pruning ratio ranging from 0.1 to 0.8 in step of 0.1. `L1 (Backbone)' means only the first convolutional layer in the backbone is pruned, and `Backbone + Neck + Head' means all the convolutional layers in backbone, neck and head are pruned with the same pruning ratio. The highest precision is marked with stars.}
\label{tab:Impact_of_ratios}
 \resizebox{3.4in}{0.65in}{
\begin{tabular}{@{}ccccccccc@{}}
\toprule
      Model                & 0.1  & 0.2  & 0.3   & 0.4   & 0.5          &0.6    &0.7   &0.8  \\ \midrule
L1 (Backbone)          & 79.2 & 79.0         & \textbf{79.8} & 79.1          & 78.6  &74.7&76.5&76.3 
       \\
L2 (Backbone)          & 77.3 & 79.0                & 78.4          & \textbf{81.1} & 80.4          &78.5&78.7&74.0\\
L3 (Backbone)          & 77.6 & 78.6                & 78.6          & 76.0          & \textbf{79.7} &78.2&78.1&79.1\\
L4 (Backbone)          & 79.0 & \textbf{81.7*} & 80.0          & 78.6          & 81.0   &   79.3 &78.6 &78.5    \\
L5 (Backbone)          & 78.5 & \textbf{80.1}       & 79.9          & 78.3          & 78.8  &77.5&78.2&77.9        \\
L1 (Head)             & 77.5 &77.9       & 77.5          &77.9 & 76.9  &\textbf{78.9}   &77.7   &77.3        \\
L2 (Head)             & 79.4 & 78.9                & 78.7          & 79.8 & 78.1 &77.1   &\textbf{80.6}   &76.7         \\
L3 (Head)             & 79.3 & \textbf{81.7*} & 77.3          & 81.7          & 79.0   &74.8&76.6&76.3       \\
Backbone + Neck + Head & 79.6 & 80.0                & \textbf{81.0} & 79.5          & 77.6          &78.6&77.9&76.4\\ \bottomrule
\end{tabular}
}
\end{table}

\noindent\textbf{Impact of pruning ratios:} To see how the pruning ratios affect the precision of P-SiamFC++, we trained P-SiamFC++ with different pruning ratios, including layer-wise ratios and global ratios. In layer-wise manner, a certain layer of SiamFC++ is pruned with a pruning ration ranging from 0.1 to 0.8 while other layers stay untouched. Due to page restrictions the neck is not considered in layer-wise manner and the same layer of the two branches in the head is pruned with the same ratio. In global manner, each convolutional layer in backbone, neck and head is pruned with the a global ratio that ranges also from 0.1 to 0.8. The precisions of P-SiamFC++ with different pruning ratios are shown in Table \ref{tab:Impact_of_ratios}. As can be seen, in layer-wise manner, the best precision is most often achieved when the pruning ratio is 0.2 while the best in the global manner happens when the pruning ratio is 0.3, which suggests that filter pruning is not only good for simplifying the model and raising efficiency but also benefit the precision, because it can improve the model's generalization ability.

\section{Conclusion}

In this work, we are the first to use rank-based filter pruning to narrow the gap between DCF- and deep learning- based trackers in UAV tracking. The proposed P-SiamFC++ achieves a remarkable balance between efficiency and precision, and demonstrates state-of-the-art performance on four UAV benchmarks. The proposed method can not only improve efficiency but, surprisingly, also improve tracking precision. We believe our work will draw more attention to model compression in UAV tracking.

\section{acknowledgement}
\footnotesize{ Thanks to the supports by Guangxi Key Laboratory of Embedded Technology and Intelligent System, Research Institute of Trustworthy Autonomous Systems, the National Natural Science Foundation of China (No. 62176170, 62066042, 61971005), the Science and Technology Department of Tibet (No. XZ202102YD0018C), and Sichuan Province Key Research and Development Project (No. 2020YJ0282).
}

\small
\bibliographystyle{ieeetr}
\bibliography{icme2022template}

\end{document}